\documentclass[a4paper,conference]{IEEEtran}
\usepackage[adjust, compress]{cite}
\usepackage{graphicx}

\usepackage{amsmath} 
\usepackage{array}
\usepackage{multirow}
\usepackage{booktabs}       

\usepackage[caption=false,font=normalsize,labelfont=sf,textfont=sf]{subfig}

\begin{document}
%
\title{A Low-Complexity Approach to Rate-Distortion Optimized Variable Bit-Rate Compression for Split DNN Computing}

\author{\IEEEauthorblockN{Parual Datta}
\IEEEauthorblockA{Intel Labs\\
Bangalore, India \\
Email: parual.datta@intel.com
}
\and
\IEEEauthorblockN{Nilesh Ahuja}
\IEEEauthorblockA{Intel Labs\\
Santa Clara, USA\\
Email: nilesh.ahuja@intel.com}
\and
\IEEEauthorblockN{V. Srinivasa Somayazulu, Omesh Tickoo}
\IEEEauthorblockA{Intel Labs\\
Hillsboro, USA\\
Emails: v.srinivasa.somayazulu@intel.com \\ 
omesh.tickoo@intel.com}
}


%


\maketitle

\begin{abstract}
Split computing has emerged as a recent paradigm for implementation of DNN-based AI workloads, wherein a DNN model is split into two parts, one of which is executed on a mobile/client device and the other on an edge-server (or cloud). Data compression is applied to the intermediate tensor from the DNN that needs to be transmitted, addressing the challenge of optimizing the rate-accuracy-complexity trade-off.  Existing split-computing approaches adopt ML-based data compression, but require that the parameters of either the entire DNN model, or a significant portion of it, be retrained for different compression levels. This incurs a high computational and storage burden: training a full DNN model from scratch is computationally demanding, maintaining multiple copies of the DNN parameters increases storage requirements, and switching the full set of weights during inference increases memory bandwidth. In this paper, we present an approach that addresses all these challenges. It involves the systematic design and training of bottleneck units - simple, low-cost neural networks - that can be inserted at the point of split. Our approach is remarkably lightweight, both during training and inference, highly effective and achieves excellent rate-distortion performance at a small fraction of the compute and storage overhead compared to existing methods.

\end{abstract}


%
\IEEEpeerreviewmaketitle

\section{Introduction}
\label{sec:Intro}

\begin{figure}
\centering
\subfloat[]{\includegraphics[width=0.48\textwidth]{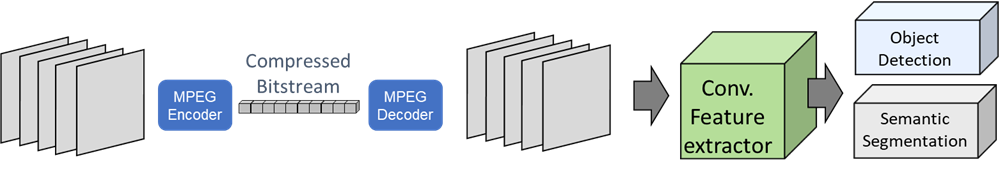}%
\label{fig:traditional}}
\hfil
\subfloat[]{\includegraphics[width=0.48\textwidth]{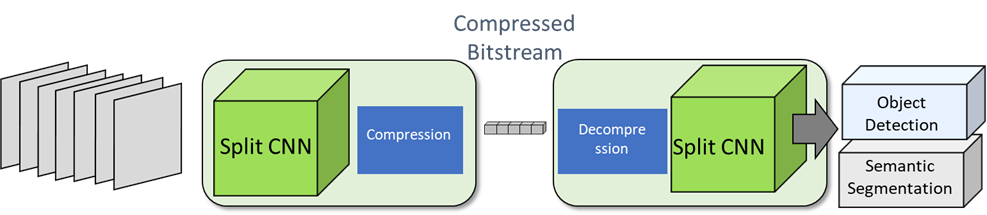}%
\label{fig:nextgen}}
\caption{Media analytics pipeline (a) Traditional, based on standards-based compression (b) Next generation based on ML-based compression and split computing}
\label{fig:mediapipe}
\end{figure}

A significant amount of visual data being generated in visual IoT applications is intended for consumption by  analytic algorithms such as classification, object detection/tracking, semantic segmentation, etc. In a typical framework, video data captured by a mobile device is compressed using conventional image and video codecs like JPEG, H.264, HEVC, etc. and transmitted over the network to an edge-server (or cloud). The bitstreams are first decompressed and then provided as input for various analytics tasks (Fig. \ref{fig:traditional}), which are increasingly being performed by very complex deep neural networks (DNNs). The scale of these IoT scenarios creates unique challenges as large numbers of IoT devices with limited compute resources rely on shared edge/cloud compute, and access to that is via a shared, time-varying network resource. This places significant new demands on visual compression algorithms to scale beyond the compression efficiencies obtained with conventional video compression standards such as H.264/HEVC. These standard compression techniques optimize perceptual quality for human consumption, which may not necessarily be optimal for analytic tasks that rely on the semantics of content rather than its perceptual quality. Consequently, task performance can degrade severely in the presence of even relatively mild compression artifacts. 

To address this, recent approaches rely on machine-learning models to learn features that are jointly optimized both for performance on an analytic task and for compression. Learnt representations from the front end or \textit{head} of a DNN -- consisting of an input layer and a number of subsequent layers -- are compressed and transmitted to an edge server. The remaining layers (also referred to as the \textit{tail}) then operate directly on these compressed representations  (Fig. \ref{fig:nextgen}). Computationally, this is much more efficient than first reconstructing the image or video input and then performing the analytic task. Early approaches \cite{chen2019toward} explored the use of simple lossless and lossy compression techniques to compress the intermediate features. While lossless techniques result in only mild reduction in bandwidth, naively quantizing the intermediate features in lossy compression leads to a drop in the task performance. Inspired by the impressive results of ML-based image compression approaches \cite{balle2016end,balle2018variational, toderici2015variable}, subsequent approaches \cite{theis2017lossy, choi2018deep, patwa2020semantic} include the quantization operation in the end-to-end training of the model. 
These approaches yield significantly better rate-distortion performance (task accuracy vs compression level) than that obtained by traditional image compression methods such as JPEG and the more recent HEIC. 

Early approaches \cite{kang2017neurosurgeon} also attempted to determine an optimal splitting point between the device and the edge-server to minimize latency. These often result in the splitting point being placed towards the end of the model, which ends up placing most of the computational burden on a low-powered, compute limited mobile device. More recent contributions \cite{eshratifar2019bottlenet, shao2020bottlenet++, matsubara2020head, matsubara2021neural} attempt to remedy this by introducing \emph{bottleneck layers} at the split point. These are layers with small number of parameters and low computational overhead which reduce the dimensions of the features to be compressed and transmitted. This allows the model to be split at earlier layers. Crucially, though, in all these approaches, once the split point has been fixed, the model is then trained for that particular partition only. Changing the split point, therefore, cannot be done without retraining the entire model, or at least a significant portion of it if techniques such as head-network distillation \cite{matsubara2020head} are used. This is a serious limitation that adversely impacts the applicability of the solution in conditions where dynamic partitioning of the workload between the mobile device and the cloud is required in response to changing compute and platform requirements at either end. Such capabilities are especially important when the edge-server needs to perform dynamic orchestration of workloads while servicing multiple mobile or client devices.

Another equally serious limitation is that the parameters of the trained pipeline are optimized for operation only at a certain compression level or bit-rate. Supporting different compression levels needed for variable bit-rate operation, therefore, also requires retraining the DNN and storing multiple copies of its weights, similar to the changing of the split point. Such multiple end-to-end trainings increase the overall training time and complexity. Moreover, during actual operation in inference mode, the entire set of parameters will have to be reloaded each time a different compression level or split point is desired. DNN parameters often run into several tens of millions, and hence storing and loading multiple copies of these parameters during runtime is expensive and can slow down overall operation. Some contributions avoid this problem by maintaining a learnable vector of scale parameters \cite{theis2017lossy} or gain parameters \cite{cui2020g} that are applied to the output of the encoder of a deep-autoencoder. While this is effective, its applicability has been demonstrated only at the output of the encoder network of a deep autoencoder, where the output dimension is already the lowest. It does not address the issue of high-dimensionality of intermediate features that arises in split computing and hence is not applicable in those scenarios.

\subsubsection*{Contributions} 
In this work, therefore, we propose a solution that overcomes all these limitations and enables variable bit-rate compression for the distributed media-analytics pipeline in Fig. \ref{fig:nextgen}. Our solution also employs bottleneck layers for feature compression. However, in contrast to other published approaches, we train  the weights of only the bottleneck layer instead of the entire network (or its head or tail portions). Further, we present an approach to designing the architecture of bottleneck. To the best of our knowledge, existing works have only explored the impact of varying some of the bottleneck layers parameters (such as number of channels), but none of them have proposed a systematic approach that results in a low-cost, yet optimal bottleneck layer design. The outcome of our approach is that only low-complexity bottleneck layers - which have far fewer learnable parameters compared to the original DNN - need to be retrained for different split points or compression levels. This reduces the overall training complexity dramatically. We demonstrate in Section \ref{sec:results} that despite its apparent simplicity, our approach outperforms a variety of benchmarks on both image classification and semantic segmentation tasks.
More importantly, it enables efficient variable bit-rate compression within the paradigm of split computing as only the parameters of a small, low-complexity bottleneck layer have to be reloaded for different compression levels.




\section{Preliminaries}
\label{sec:Preliminaries}
\begin{figure}
	\centering
	\includegraphics[width=0.45\textwidth]{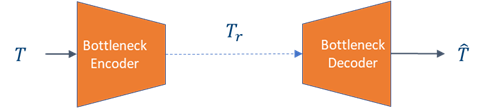}
	\caption{Bottleneck unit.}
	\label{fig:bottleneckunit}
\end{figure}

Here, we provide some background information on intermediate feature compression and training DNNs for split computing. As explained earlier, we split a DNN model that has been designed for a particular task like classification or segmentation into a front-end or head-network and a back-end or tail-network as shown in Fig. \ref{fig:nextgen}. The intermediate features are to be transmitted to the tail that resides on an edge-server or cloud. The dimensions of the intermediate features, especially towards the earlier layers of the DNN, tend to be very high making them unsuitable for efficient compression. Therefore, the use of bottleneck units has been proposed. The unit comprises two modules –- a Bottleneck Encoder and a Bottleneck Decoder –- as shown in Fig. \ref{fig:bottleneckunit}. The bottleneck unit can hence be viewed as a deep autoencoder that has been designed for the feature space of a DNN rather than for its input. The Bottleneck Encoder transforms the high-dimensional intermediate feature tensor into an appropriate lower dimensional space. This lower dimensional tensor is compressed and transmitted to the edge-server where the Bottleneck Decoder restores it to its original dimension.

Early works explored the use of lossless of simple lossless compression techniques for intermediate DNN features \cite{chen2019toward}, but these don't yield a significant reduction in bandwidth nor offer flexibility in adapting the compression bitrate. For this reason, lossy compression is applied, which involves first quantizing the intermediate features by some quantization step-size, $Q$, followed by lossless compression of the resulting discrete-valued signal by algorithms such as Huffman Coding or Arithmetic Coding. Simply quantizing the intermediate features during inference, however, leads to a drop in the task performance. To overcome this, the quantization operation is included in the training of the network. Since quantization is a non-differentiable operation, it cannot be directly incorporated during a gradient-based training regimen. This can be circumvented by either replacing quantization by additive noise \cite{balle2016end}, or using a stochastic form of binarization \cite{toderici2015variable}, or using a smooth approximation of the gradient \cite{theis2017lossy}. In this work, we adopt the third approach \cite{theis2017lossy}, using an identity function as an approximation for quantization during backpropagation. Note that this approximation is applied only to the gradients during backpropagation, and during the forward pass, true quantization is performed.




The model itself is trained with a loss function of the form:
\begin{equation}
\label{eq:training_loss}
    L = L_r + \alpha L_t,
\end{equation}
where, $L_t$ is a task-loss to maximize task performance, $L_r$ is a rate-loss term to minimize bit-rate of the encoded data, and $\alpha$ is a term that controls the relative weighting of the two terms. Higher values of $\alpha$ assign higher weight to $L_t$ relative to $L_r$, i.e. emphasize task-accuracy more than bit-rate, resulting in higher accuracy but at higher bit-rates. Conversely, lower values of $\alpha$ allow the network to achieve lower bit-rates, but at the cost of task performance. Note that both $L_r$ and $L_t$ are functions of the quantization step-size, $Q$, since true quantization is performed in the forward pass during training and this directly affects the values of the loss terms. 

For $L_t$, the usual loss functions such as cross-entropy loss for classification, and sum of per-pixel cross-entropy losses for segmentation are used. Various approaches have been explored in literature to choose an appropriate $L_r$. Since the feature values are discrete-valued following quantization, the rate is the expected code length which is lower-bounded by the entropy of the probability distribution of the quantized alphabet \cite{cover1999elements}. In \cite{theis2017lossy}, a continuous, differentiable function is used to approximate this probability distribution. In \cite{balle2016end, balle2018variational}, methods of variational inference are used to approximate the true distribution by parametric variational densities. Other contributions in literature adopt indirect approaches. In \cite{cheng2018deep}, the $\ell_2$ norm of the compressed feature is taken as a proxy for the number of bits required to encode the data. In \cite{liu2018cnn}, the $\ell_1$ norm of the DCT coefficients of the original image was used as an estimate for rate. \cite{alvar2019multi} improved upon this by incorporate spatial prediction into the calculation of the loss to get a better estimate of the rate. In our work, we adopt the indirect approach and use the $\ell_1$-norm of the output of the bottleneck encoder as a proxy for the rate. As will be demonstrated shortly, even this simple approach yields impressive results.




\begin{figure}
	\centering
	\includegraphics[width=0.45\textwidth]{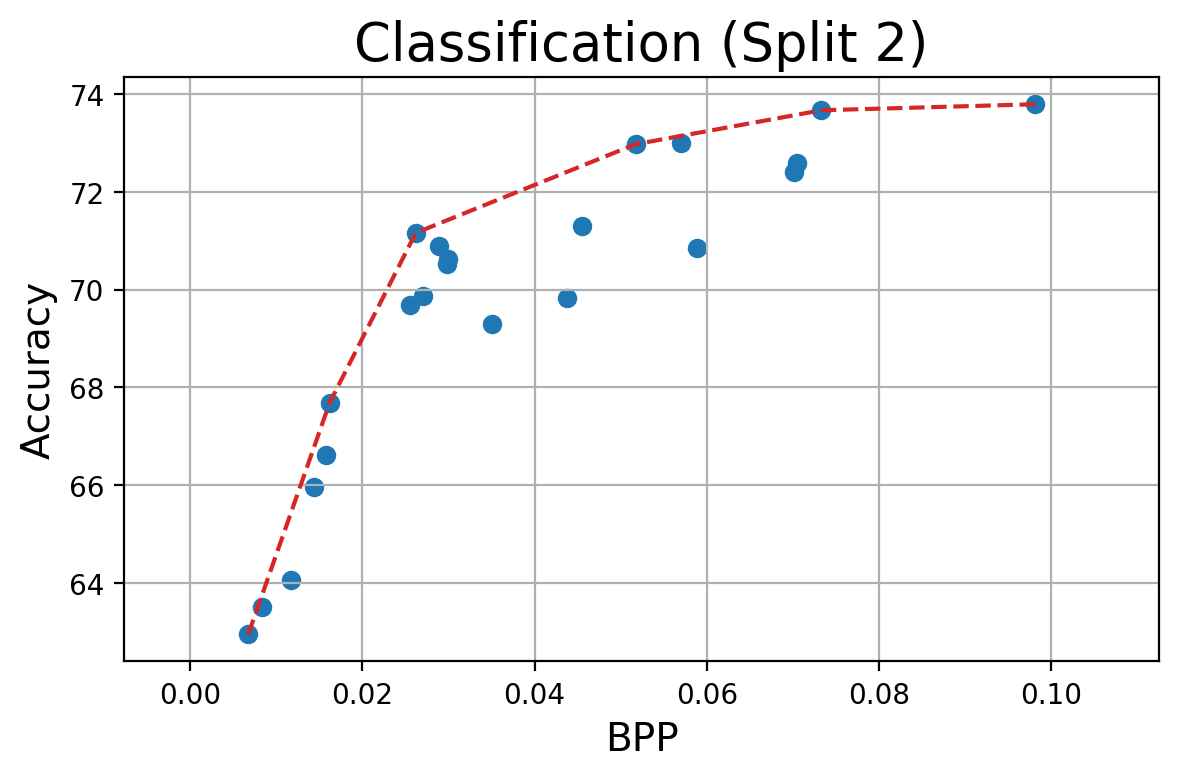}
	\caption{Scatter plot of (accuracy, bpp) pairs and Pareto frontier of the same.}
	\label{fig:parametersearch}
\end{figure}

\section{Method}
\label{sec: Method}

In this section, we describe a systematic approach to the design and training of low-complexity bottleneck layers for flexible workload partitioning and enabling variable bit-rate compression in the split computing paradigm.
\subsection{Design of Bottleneck Unit}

We consider a split DNN architecture, where the output of the last layer on the client side is an intermediate layer of the DNN, a tensor with dimension $H\times W \times C$. 
The Bottleneck Encoder as shown in Fig. \ref{fig:bottleneckunit} transforms this high-dimensional vector into an appropriate lower dimensional space, $H_r\times W_r \times C_r$ such that $H_r \leq H,W_r \leq W,C_r \leq C$, where $H_r,W_r,C_r$ are parameters yet to be determined. The Bottleneck Decoder, which restores the feature to its original dimension is a mirror image of the Encoder. 

Next, we must choose an appropriate topology for the bottleneck encoder. The Bottleneck Encoder can be designed as a single-layer or a multi-layer network, with considerable flexibility in the choice of each layer’s topology such as fully-connected layers, convolutional layers, depthwise separable convolutional layers, residual layers, etc. Importantly, we wish to keep the overhead introduced by the additional bottleneck processing low relative to the overall compute -- both in terms of number of additional parameters and number of additional flops. For our initial study, therefore, we choose to model bottleneck units by depthwise separable convolutional layers as these have the fewest parameters and the lowest computational complexity \cite{howard2017mobilenets}. We note that use of other relatively more complex architectures could potentially provide better results. Finally, we need to choose the following hyperparameters: convolutional kernel size, output channels $C_r$, and output resolution $(H_r, W_r)$. We fix the kernel size to $3\times 3$. The output resolution is related to the input by a stride value $S$, which effectively results in a downsampling by $S$, i.e. $H_r=H/S$ and $W_r=W/S$. Thus, there are two parameters, $C_r$ and $S$, over which we have to search for an optimal architecture.

\begin{table}
	\centering
	\caption{Parameters and their ranges for search space exploration.}
	\label{table:searchspaceparams}
	\begin{tabular}{ccc}
		\toprule
		\multirow{3}{*}{Parameter}  & \multicolumn{2}{c}{Range}  \\
		\cmidrule(r){2-3}
		  & {\begin{tabular}{c}
		     Classification  \\
		     (Resnet50) 
		\end{tabular}}  
		& {\begin{tabular}{c}
		     Segmentation  \\
		     (DeepLab v3) 
		\end{tabular}}
		\\
		\cmidrule(r){1-1}\cmidrule(r){2-2}\cmidrule(r){3-3}
		 No. of Channels $C_r$ & $\{32, 64, 96, 128\}$ & {\begin{tabular}{c}
		     $\{2, 4, 8, 16,$   \\
		     $32, 48, 64\}$ 
		\end{tabular}} \\
		\midrule
		 Stride $S$ & \multicolumn{2}{c}{$\{2,4,6\}$}\\ 
		 \midrule
		 Quant. parameter $Q$ & [1.0, 16.0] & [0.5,24.0] \\
		 \midrule
		 $L$, where $\alpha=10^L$ &  \multicolumn{2}{c}{[-4.0,-1.0]}\\
		 
		\bottomrule
	\end{tabular}
\end{table}

\subsection{Training}
\label{sec:method_training}

\begin{table*}[hbtp]
	\centering
	\caption{Split points within the models. \% Compute is the \% of the total compute till the split point.}
	\label{table:splitpoints}
	\begin{tabular}{ccccccc}
		\toprule
		 & \multicolumn{3}{c}{Resnet50}  & \multicolumn{3}{c}{DeepLab v3} \\
		\cmidrule(r){2-4}\cmidrule(r){5-7}
		 & Layer & Dimension & \% Compute & Layer & Dimension & \% Compute \\
		\cmidrule(r){1-1}\cmidrule(r){2-2}\cmidrule(r){3-3}\cmidrule(r){4-4}\cmidrule(r){5-5}\cmidrule(r){6-6}\cmidrule(r){7-7}
		 Split 1 & layer4[2]  & $2048\times 7\times 7=100352$  & 99.90 & Classifier[0].conv[1] & $256\times 65\times 65=1081600$ & 74.32\\
		 Split 2 & layer4[1]  & $2048\times 7\times 7=100352$ & 81.77 & Resnet50.layer4[2] & $2048\times 65\times 65=8652800$ & 60.99\\
		 Split 3 & layer4[0]  & $2048\times 7\times 7=100352$ & 63.55  & Resnet50.layer4[1] & $2048\times 65\times 65=8652800$ & 49.83 \\
		 Split 4 & layer3[4]  & $1024\times 14 \times 14=200704$ & 34.01  & Resnet50.layer4[1] & $1024\times 65\times 65=4326400$ & 23.43 \\
		\bottomrule
	\end{tabular}
\end{table*}

We start by using a model that has already been trained stand-alone on a particular task, i.e. without splitting the model with bottleneck layers and without compression of intermediate features. Then we insert a bottleneck unit into this pretrained model at the desired split point. The features at the output of the bottleneck encoder are then compressed and training is performed using the approach described in section \ref{sec:Preliminaries}, but with one crucial difference: only the weights of the bottleneck layers are trained; the weights of the rest of the model, which have already been pretrained, are not updated. This dramatically reduces the training time because: (a) the number of learnable parameters in bottleneck units is a small fraction of the total number of weights in the entire model (less than 1\%, see Table \ref{table:modelcomplexity}), and (b) training small bottleneck units requires far fewer training epochs (typically under 15) compared to training the entire network (typically over 50). Cumulatively, these two factors end up reducing the overall training complexity by orders of magnitude. 
One might expect that since all the parameter of the entire model are not being updated, the performance of this approach (task accuracy and compression) could suffer. Remarkably, we show in Section \ref{sec:results}, that drop in performance is actually very small. 

\subsection{Parameter Space Exploration}

We have the following set of four hyperparameters that influence the eventual rate-distortion performance of the pipeline:
\begin{itemize}
    \item Two hyperparameters relating to the bottleneck architecture: the number of channels at the output of the bottleneck decoder, $C_r$, and the stride of the convolutional kernel, $S$, as explained earlier.
    \item Two hyperparameters relating to compression: the Lagrange multiplier $\alpha$ from Eq. \ref{eq:training_loss}, and the quantization step-size, $Q$
\end{itemize}
These hyperparameters interact in complex ways. For instance, a reduction in bit-rate can be obtained by reducing $\alpha$ as explained in Section \ref{sec:Preliminaries}; or by increasing $Q$; or by reducing $C_r$ or $S$, either of which will reduce the dimension of the bottleneck encoders output. However, each of these will impact task accuracy in different ways. We therefore have to search this hyperparameter space in order to determine the set of parameters that yield optimal performance. For this, we adopt approaches from the well researched field of hyper-parameter optimization for training deep neural networks \cite{bergstra2011algorithms, bergstra2012random}. Some of the existing approaches include grid-search, random-search, and evolutionary algorithms. In our work, we adopt a random-search approach in the 4-dimensional space of $(C_r, S, \alpha, Q)$ since, for such low-dimensional spaces, random-search has been shown to yield good results at a small fraction of the computational cost of other more sophisticated methods \cite{bergstra2012random}. Table \ref{table:searchspaceparams} shows the set or range of values used for each of the hyperparameters for the two different tasks that we have experimented with. We assume a uniform probability mass prior over the discrete-valued variables $C_r$ and $S$, and a uniform probability density prior over the continuous-valued $Q$. The dynamic range of $\alpha$ values that need to be searched over is large; hence, we express $\alpha=10^L$, and impose a uniform prior over $L$.

For each such 4-tuple, we train the bottleneck layers and measure the resultant task metric (accuracy or mIOU) and rate (bpp, or bits-per-pixel). We then generate a scatter plot of the task metric against the bpp, from which we extract the Pareto frontier. The points on this frontier represent the optimal tradeoff between task performance and compression level. An example of such a plot is shown in Fig. \ref{fig:parametersearch}. We use the SigOpt software \cite{Sigopt} to perform this search space optimization.

Note that such a strategy for search space exploration, which requires training and evaluation at multiple parameter values, would be prohibitively expensive with existing approaches owing to the large training times involved. Our approach of training only the bottleneck layer weights, as described in the previous section, makes the use of such a parameter space exploration more feasible.

\section{Experiments and Results}
\label{sec:results}

\begin{table}
	\centering
	\caption{Model Complexity: Parameters and MAC counts.}
	\label{table:modelcomplexity}
	\begin{tabular}{ccc}
		\toprule
		Model & Total Parameters & Total compute (MAC) \\
		\cmidrule(r){1-1}\cmidrule(r){2-2}\cmidrule(r){3-3}
		\multicolumn{3}{c}{\textbf{Classification}} \\
		 Resnet50  & 25.55 M & 16.48 G \\
		 Bottleneck (Max) & 0.28 M & 0.06 G \\
		\midrule
		\multicolumn{3}{c}{\textbf{Segmentation}} \\
		 Deeplab v3  & 42.01 M & 134.73 G  \\
		 Bottleneck (Max) & 0.34 M & 0.40 G \\
		\bottomrule
	\end{tabular}
\end{table}

We have tested our approach on two analytic tasks: image classification and semantic segmentation. For the classification task, we test on the ImageNet dataset \cite{imagenet_cvpr09} and use a Resnet50 \cite{he2016deep} model pretrained on the same. The images are scaled to $224\times 224$ resolution prior to classification. For segmentation, we test on the validation split of the MS-COCO 2017 dataset \cite{lin2014microsoft}, and use a DeepLab v3 model \cite{chen2017rethinking} pretrained on its train split. For this task, the images are scaled to a resolution of $513 \times 513$. For both tasks, we test our approach at four different split points shown in see Table \ref{table:splitpoints} within their respective models. The layer names for each model are the same as those used in their respective implementations in Torchvision \cite{Torchvision}.

\begin{table*}
	\centering
	\caption{Sample bpp values.}
	\label{table:bppvalues}
	\begin{tabular}{ccccccccc}
		\toprule
		  & S1 & S2 & S3 & S4 & \multirow{2}{*}{Entropic Student\cite{matsubara2021neural}} & \multirow{2}{*}{Variational \cite{balle2018variational}} & \multirow{2}{*}{JPEG} & \multirow{2}{*}{HEVC}  \\
		 & \multicolumn{4}{c}{(Ours at different splits)} &  &  &  &   \\
		 \cmidrule(r){1-1}\cmidrule(r){2-5}\cmidrule(r){6-6}\cmidrule(r){7-7}\cmidrule(r){8-8}\cmidrule(r){9-9}
		 70\% classification accuracy & 0.0115 & 0.0243  & 0.0332  & 0.1041 & 0.3176 & 0.8235 & 1.0237 & 0.2949  \\
		 0.66 mIOU & 0.0028  & 0.0487  & 0.0728 & 0.1257 & 1.1078 & 1.0526 & 1.3258 & 0.8346  \\

		\bottomrule
	\end{tabular}
\end{table*}

\begin{figure}
	\centering
	\includegraphics[width=0.45\textwidth]{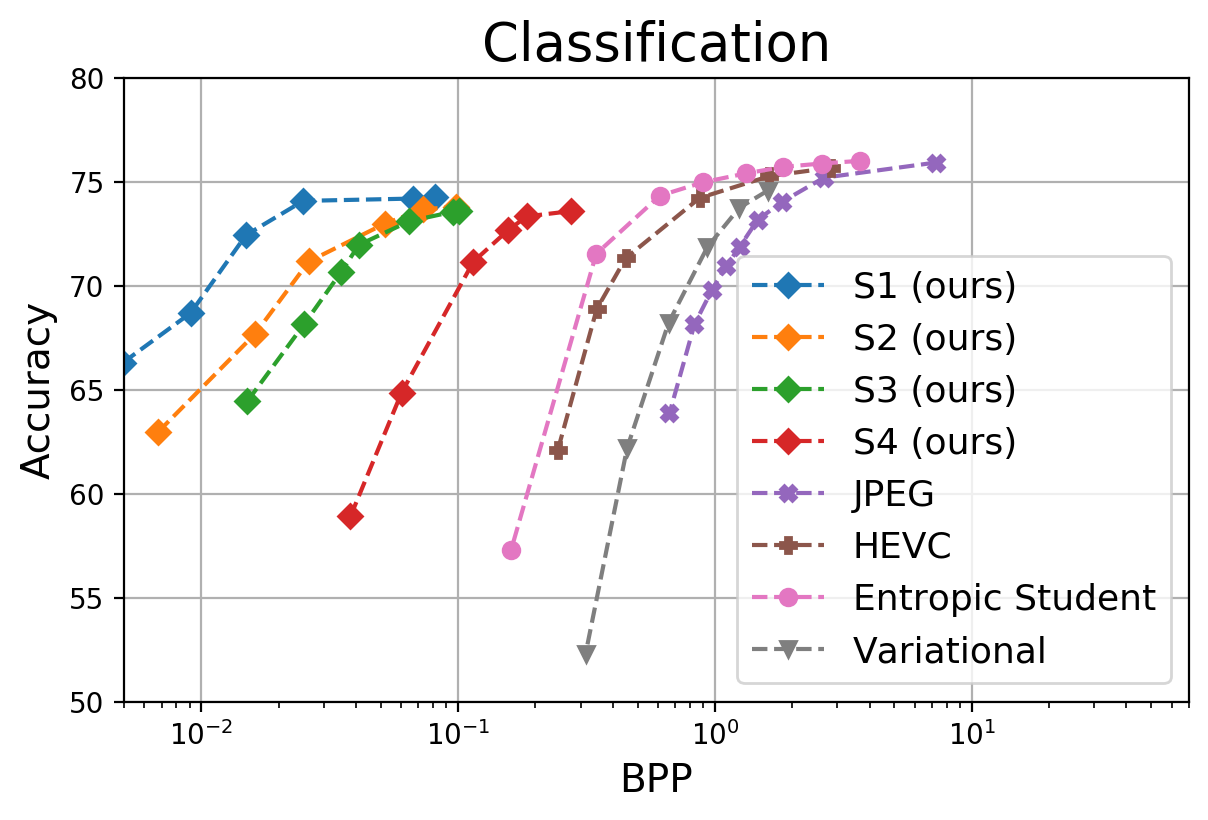}
	\caption{Accuracy vs bpp on Imagenet classification using Resnet50. Our method achieves significantly lower bit-rates ($\sim 3-15$x lower) compared to other methods at similar accuracy levels for all splits (S1 to S4), but slightly lower peak accuracy ($\sim 0.5\%-1.5\%$ lower). }
	\label{fig:classification}
\end{figure}

\begin{figure}
	\centering
	\includegraphics[width=0.45\textwidth]{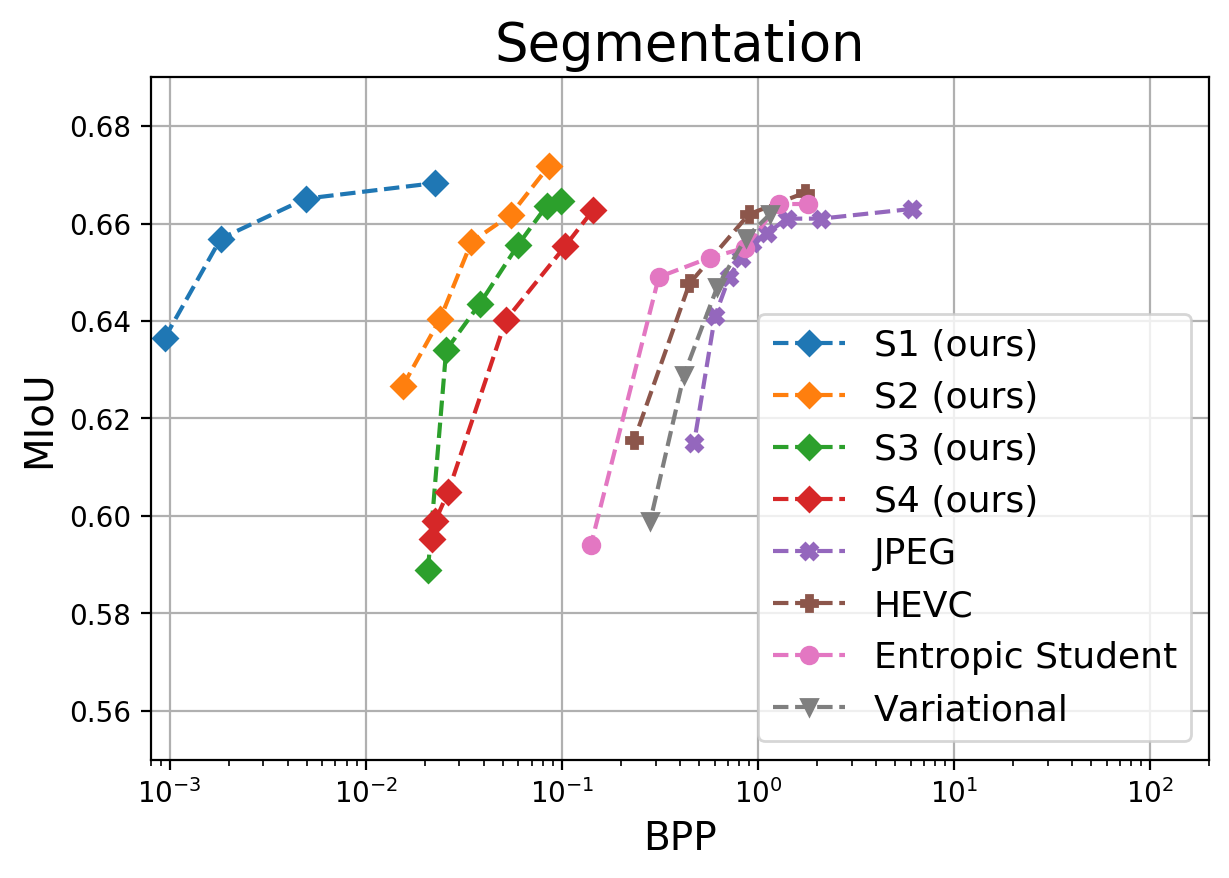}
	\caption{mIOU vs bpp on COCO2017 segmentation using Deeplab v3. Our method achieves significantly lower bit-rates ($\sim 10 - 100$x less) compared to other methods at similar mIOU levels for all splits (S1 to S4), and also slightly higher  peak mIOU ($\sim 1\%$ higher). }
	\label{fig:segmentation}
\end{figure}

We benchmark our method against standard image compression algorithms - JPEG and  HEIC (high-efficiency image compression) - and also against some of the recent ML-based image compression methods which include the Entropic Student  \cite{matsubara2021neural} and variational image compression \cite{balle2018variational}. To compare and benchmark the effectiveness of our approach, we derive the Pareto optimal accuracy-vs-compression curve, which is a plot of a task-specific accuracy metric against the compression level. For classification, the metric is simply the classification accuracy; for segmentation, it is the Jaccard index as measured by the mean intersection-over-union (mIOU). The compression level is represented by bits-per-pixel (bpp), which, as the name indicates is the number of bits of information that needs to be transmitted divided by the number of pixels in the input image. The results for \cite{matsubara2021neural} were originally reported in terms of file sizes, but we convert those into equivalent bpp values using $\text{bpp}=\text{file\_size}/(W\times H)$. The results are plotted in Fig. \ref{fig:classification} for classification and Fig. \ref{fig:segmentation} for segmentation. For our method, results at all the different split points tested are plotted in the same figure.

Our method shows significant reduction in bit-rates compared to other methods across all accuracy levels for classification, and across all mIOU levels for segmentation. This is true for all the split points that were tested. Sample bpp values at specific accuracy and mIOU levels are shown in Table \ref{table:bppvalues}. At an accuracy of $70\%$, the bit-rate of our method is 3 to 27 times lower than the best performing amongst the benchmarks that we tested. For segmentation, the results are even more impressive with our method achieving bit-rates that are 9 to 392 times lower. The peak classification accuracy with our approach is slightly lower (within 0.5\%-1.5\%) than that from other methods, but for segmentation, the peak mIOU is higher than all other methods. 

\begin{figure}
	\centering
	\includegraphics[width=0.45\textwidth]{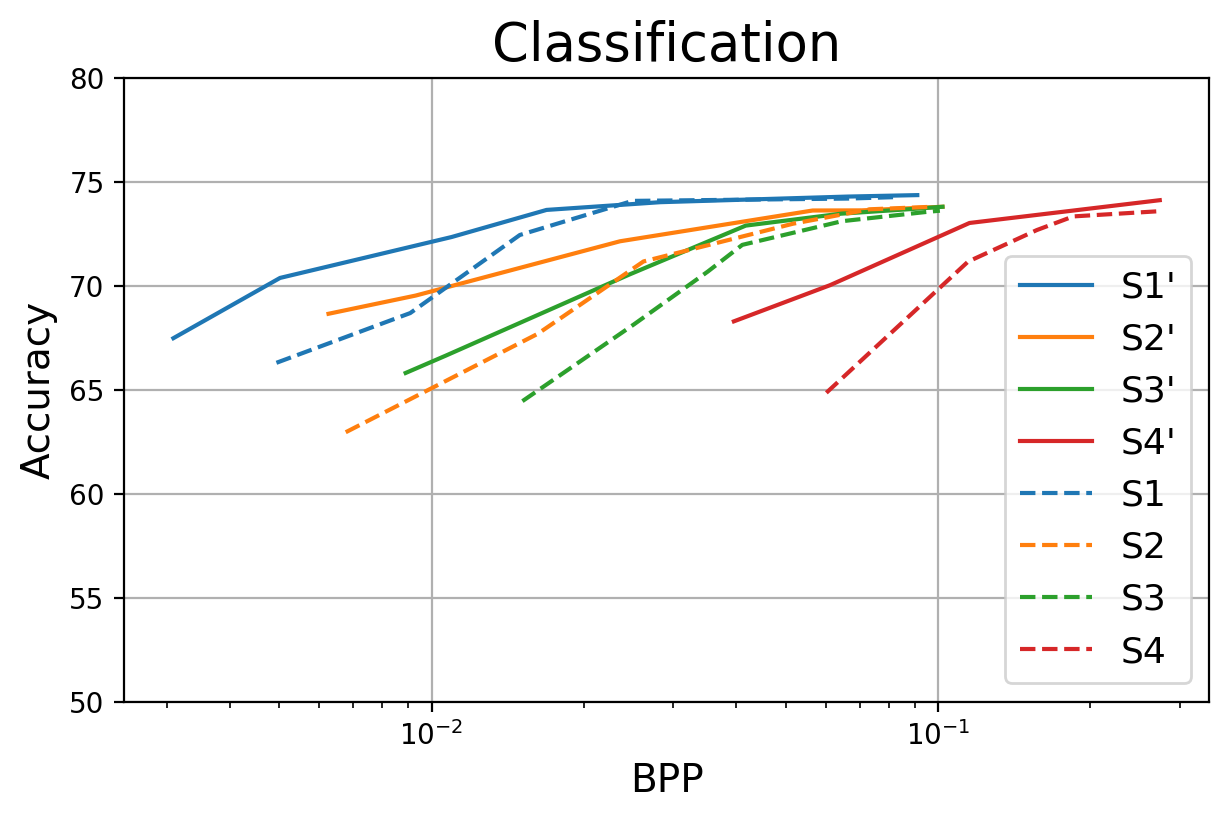}
	\caption{Accuracy vs bpp with full-model training (S1' to S4') and bottleneck only training (S1 to S4).}
	\label{fig:classification_share_vs_sep}
\end{figure}

\begin{figure}
	\centering
	\includegraphics[width=0.45\textwidth]{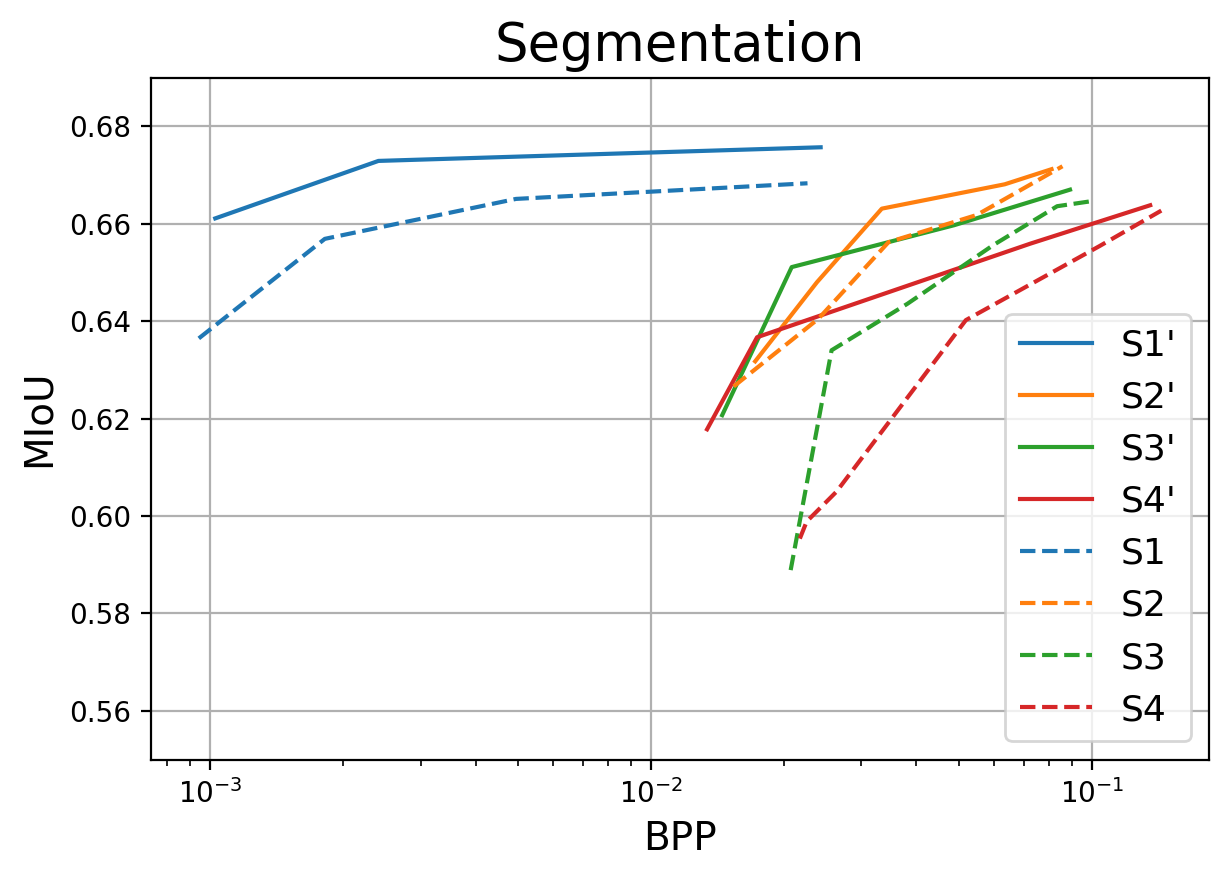}
	\caption{mIOU vs bpp with full-model training (S1' to S4') and bottleneck only training (S1 to S4).}
	\label{fig:segmentation_share_vs_sep}
\end{figure}

Note that each point on the Pareto optimal performance curves of our method represents a different point in the 4-dimensional parameter space $(C_r, S, \alpha, Q)$. This implies that the size (number of learnable weights) and computational complexity of the bottleneck layer, which depend on $C_r$ and $S$, differs for each point. In Table \ref{table:modelcomplexity}, therefore, we report the size and complexity of the largest bottleneck unit resulting from our search procedure. As shown, even this largest bottleneck layer is a negligibly small fraction of the size and complexity of the overall model. In the supplementary material, we document in detail the 4-d parameter values for each point on the Pareto optimal curve for all splits. One point to note in a variable bit-rate setting is that the information about which hyperparameter set has been used for bottleneck encoding must be sent over to the edge-server, so that the corresponding correct bottleneck decoder can be used. The impact to bit-rate, however, is vanishingly small. In our results, for example, a maximum of 7 different hyperparameter sets have been used to generate a Pareto optimal performance curve, which implies a maximum of 3-bits ($\lceil \log_2 7 \rceil=3)$ of additional overhead, which translates to a bpp increase of less than $6\times 10^{-5}$.

Finally, we evaluate the performance of the method if \emph{all} the parameters of the pipeline - both the original model and the bottleneck layers - are trained rather than only the bottleneck parameters. Recall that not only does this increase the number of parameters to be trained by over a 100-fold (from Table \ref{table:modelcomplexity}), it also requires more epochs to train (as explained in section \ref{sec:method_training}). The results are shown in Figures \ref{fig:classification_share_vs_sep} and \ref{fig:segmentation_share_vs_sep} for classification and segmentation, respectively. The performance curves for full-model training are shown in solid lines, while the curves for bottleneck-only training are shown in dashed lines. The full-model curves show mild improvement over the bottleneck-only curves, but this comes at the cost of dramatically higher training time. Equally importantly, as explained in Section \ref{sec:Intro}, our method enables practical variable bit-rate compression, as only the parameters of the bottleneck layers have to be reloaded to switch to a different compression level, and these are a small fraction (less than 1\%, see Table \ref{table:modelcomplexity}) of the number of parameters in the original model.

\section{Conclusions and Future Work}
In this paper, we have presented an approach that enables flexible workload partitioning and enables practical variable bit-rate compression in distributed media analytics pipeline. Our approach is remarkably lightweight, both during training and inference, highly effective and achieves excellent rate-distortion performance at a small fraction of the compute and storage overhead compared to existing methods.

Finally, we note that in this initial study of ours, we have intentionally made simple, low-complexity choices for various parts of our method. These include the architecture of bottleneck layers (single depthwise separable convolutional layer); the rate-loss term ($\ell_1$-norm of the intermediate feature); and for the compression scheme (simple uniform quantization followed by Huffman encoding). For each of these, there are more sophisticated alternatives that could help improve the performance further, such as increasing the number of layers in the bottleneck units or using different topologies for the bottleneck layers; using arithmetic coding during lossless compression; and using entropy estimation for the rate-loss term during training. Hence, although our approach already yields impressive bit-rate reduction over state-of-the-art methods, we believe that there is still room for further improvement via exploring the approaches just outlined for future work. 

\bibliographystyle{IEEEtran}
%
\bibliography{IEEEabrv, Bottleneck}

\end{document}